\documentclass{article}

\usepackage[preprint]{neurips_2026}

\usepackage[utf8]{inputenc}
\usepackage[T1]{fontenc}
\usepackage{hyperref}
\usepackage{url}
\usepackage{amsfonts}
\usepackage{amsmath}
\usepackage{amssymb}
\usepackage{nicefrac}
\usepackage{microtype}
\usepackage{graphicx}
\usepackage{xcolor}
\usepackage{booktabs}
\usepackage{multirow}
\usepackage{xcolor,colortbl}
\usepackage{wrapfig}
\usepackage{pifont}
\usepackage{enumitem}
\usepackage{caption}
\usepackage{capt-of}
\usepackage{placeins}
\usepackage[table]{xcolor}

\definecolor{GroupGray}{RGB}{246,247,249}
\definecolor{OursGray}{RGB}{238,241,245}

\newcommand{\cmark}{\ding{51}}
\newcommand{\xmark}{\ding{55}}
\newcommand{\best}[1]{\textbf{#1}}

\title{Touch-R1: Reinforcing Touch Reasoning in MLLMs}

\author{%
Yingxin Lai$^{1,5}$\thanks{Equal contribution. Author order is random.} \quad
Yafei Zhou$^{2,5}$\footnotemark[1] \quad
Fucai Zhu$^{5}$ \quad
Siyu Zhu$^{3}$ \quad
Weihao Yuan$^{4,5}$\thanks{Corresponding author.} \\
$^{1}$Xiamen University \quad
$^{2}$Great Bay University \quad
$^{3}$Fudan University \quad
$^{4}$Nanjing University \\
$^{5}$Daimon Robotics
}

\begin{document}

\maketitle

\begin{abstract}
While rule-based reinforcement learning has recently catalyzed explicit reasoning in multimodal models, tactile reasoning remains largely underexplored. Existing tactile-language models primarily rely on supervised or contrastive objectives, which limits their capacity to ground predictions in physical evidence or rectify misleading visual priors. Tactile reasoning introduces two modality-specific challenges: the ordinal nature of physical attributes (e.g., hardness, roughness) and the cross-sensor distribution shifts inherent in optical tactile hardware. In this work, we introduce TouchReason-1M, a large-scale multimodal dataset comprising over 1M synchronized tactile pairs across four distinct sensors, and TouchReason-Bench, a rigorous framework for evaluating tactile perception and visual-tactile conflict resolution. Building upon these, we propose Touch-R1, a tactile reasoning MLLM based on Qwen2.5-VL-7B. Touch-R1 is trained via a tactile-grounded GRPO objective that combines ordinal-aware accuracy, cross-sensor physical consistency, structured-format control, and an input-side tactile grounding objective. Specifically, the tactile-use reward assigns credit only when authentic tactile inputs yield superior correctness relative to counterfactual controls where the tactile stream is removed, shuffled, or noise-masked. On TouchReason-Bench, Touch-R1-7B outperforms Octopi-13B by 18.4\% and GPT-4o by 24.7\% on average. Its structured reasoning traces reveal emergent behaviors of probing, comparison, and revision, demonstrating that R1-style reasoning can be effectively grounded in physical contact. Our code and data will be made public on the \href{https://laiyingxin2.github.io/Projects/}{project page}.
\end{abstract}

\section{Introduction}

Humans rarely infer physical properties from vision alone. A visual impression is often followed by tactile contact, and the resulting physical evidence may revise the initial judgment when appearance is misleading. This perception-verification-revision loop is central to embodied interaction with the physical world. Yet current multimodal LLMs often fail on precisely this loop: in visual-tactile conflict settings, models such as GPT-4o and Gemini-2.5-Pro tend to follow the visual object prior rather than revise their predictions from tactile evidence. Our experiments confirm this gap on TouchReason-Bench, where general-purpose MLLMs substantially underperform Touch-R1, and qualitative examples show that tactile-grounded reasoning can correct visually plausible but physically inconsistent predictions. This suggests a paradigm gap: current models can associate touch with language, but do not reliably use touch as evidence for physical-property revision. In this work, we study tactile-grounded reasoning. For the tactile modality, we focus on optical tactile sensors because they provide high-resolution contact observations for tactile reasoning.

\begin{figure}[t]
\centering
\vspace{-7mm}
\resizebox{\linewidth}{!}{\includegraphics{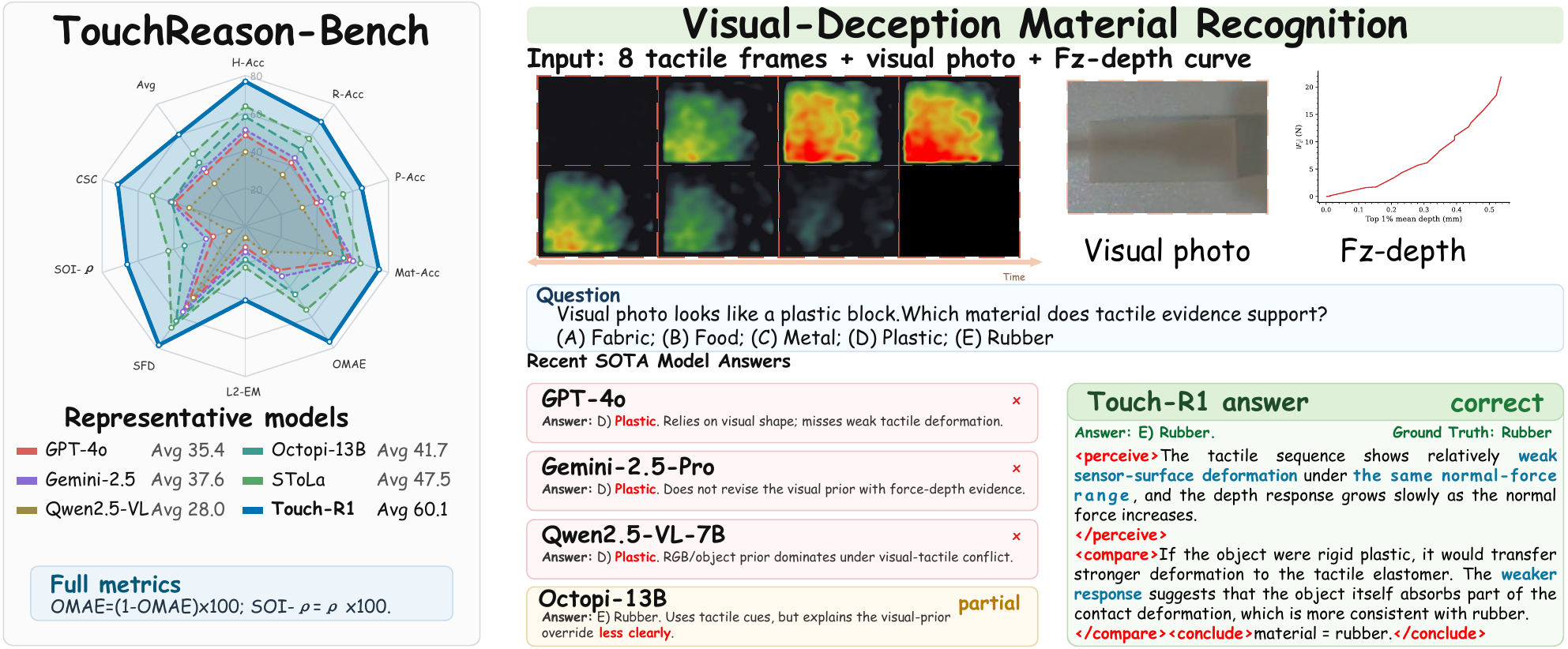}}
\vspace{-6mm}
\caption{Motivating example and benchmark overview. Left: Touch-R1 improves over representative closed-source MLLMs, open-source VLMs, and tactile-specialist models on TouchReason-Bench. Right: in a visual-deception material-recognition example, the visual photo suggests plastic, while tactile frames and the force-depth curve indicate compliant deformation. Touch-R1 revises the visual prior and predicts rubber using structured tactile-grounded reasoning.}
\label{fig:laser}
\vspace{-6mm}
\end{figure} 

Recent tactile-language models, including Octopi~\cite{yu2024octopi}, TVL~\cite{fu2024tvl}, Touch100k~\cite{cheng2025touch100k}, AnyTouch~\cite{feng2025anytouch}, SToLa~\cite{cheng2026stola}, and VitaTouch~\cite{zong2026vitatouch}, have advanced tactile-language modeling through supervised fine-tuning, contrastive alignment, and mixture-of-experts architectures. However, these methods do not 
verify whether the final answer is grounded in tactile evidence. As a result, they may learn visually plausible or object-category-driven answers without learning when tactile evidence should override vision.
In this work, we introduce rule-based RL objectives to establish tactile-use verification.

The R1 paradigm \cite{shao2024deepseekmath} has recently been extended to vision \cite{huang2025vision,shen2025vlm,liu2025visual} and video \cite{feng2025video,chen2025grpo}, but these extensions largely rely on strong pretrained priors in their input modalities. Optical tactile imagery violates this assumption in three ways. First, marker displacement, elastomer deformation, and contact-induced texture patterns are largely absent from natural-image pretraining corpora. Second, GelSight Mini~\cite{yuan2017gelsight}, Xense~\cite{xense2025photon}, Tac3D~\cite{zhang2022tac3d}, DM-Tac X~\cite{daimon_dmtacx_2026}, and other optical tactile sensors differ in resolution, marker layout, illumination, and optical geometry, making pixel-level alignment across sensors unreliable. Third, physical attributes are ordinal rather than categorical: predicting moderate hardness for a soft object is physically closer than predicting hard, but sparse binary rewards treat both mistakes equally. These properties make tactile reasoning a poor fit for standard supervised alignment and vanilla rule-based RL.

 To address these problems, we propose Touch-R1, a tactile reasoning MLLM trained through three stages: tactile dynamics pretraining, QA supervised fine-tuning, and rule-based reinforcement learning. First, a ViT-based tactile encoder is trained to predict the latent code of the next tactile frame, encouraging sensitivity to deformation dynamics rather than static appearance. Second, we construct TouchReason-1M, a four-optical-sensor tactile reasoning dataset with over 1M synchronized tactile data pairs, and use verified reasoning-style QA annotations to align tactile observations with structured physical-property answers. Third, we train Touch-R1 with a tactile-grounded GRPO objective that evaluates final conclusions against ground-truth labels and assigns tactile-use credit only when real tactile inputs improve correctness over missing, temporally shuffled, or sensor-matched noise-masked controls. This design prevents the policy from receiving tactile-use credit for answers that can be solved from visual or object-category priors alone.

Touch-R1 combines tactile-specific output rewards with an input-side tactile grounding objective. The Ordinal-Aware Accuracy Reward replaces binary correctness with dense credit over ordered attributes such as hardness, roughness, and protrusion, so adjacent-level errors are penalized less than larger ordinal deviations. The Cross-Sensor Physical Consistency Reward encourages paired sensors observing the same object to agree in physical-property space without requiring pixel-level sensor alignment. The Structured-Format Reward stabilizes the \texttt{<perceive>}, \texttt{<compare>}, and \texttt{<conclude>} interface required for reward parsing. In addition, the input-side tactile grounding objective perturbs tactile streams and encourages the policy distribution to depend on tactile evidence rather than visual or object-category shortcuts. In experiments, these rewards reduce ordinal error, improve cross-sensor consistency, and help the model revise misleading visual priors using tactile evidence. Our contributions can then be summarized as follows:
\begin{itemize}[leftmargin=*]
    \item We propose \textbf{Touch-R1}, a tactile reasoning MLLM trained with tactile-grounded GRPO, combining ordinal-aware accuracy, cross-sensor physical consistency, structured-format control, and input-side tactile grounding to encourage physically grounded tactile reasoning.
    \item We construct \textbf{TouchReason-1M}, a multimodal tactile reasoning dataset containing over 1M synchronized tactile data pairs collected with four optical tactile sensors, and introduce \textbf{TouchReason-Bench} for evaluating tactile perception, ordinal comparison, cross-sensor consistency, and visual-tactile conflict.
    \item Extensive experiments show that Touch-R1 improves tactile reasoning accuracy, reduces ordinal error, strengthens cross-sensor consistency, and produces structured perception-comparison-revision rationales grounded in physical contact.
\end{itemize} 
\section{Related Work}
\vspace{-2mm}

\noindent \textbf{Tactile perception and tactile-language models.}
Optical tactile sensors~\cite{yuan2017gelsight,lambeta2020digit,zhang2022tac3d,daimon_dmtacx_2026}
share an elastomer-plus-camera design, but differ in resolution, marker layout,
illumination, and optical geometry. This heterogeneity has motivated
cross-sensor tactile representation learning~\cite{feng2025anytouch,yang2024unitouch,zhao2024transferable,higuera2024sparsh}.
Recent tactile-language models connect touch to language through supervised
fine-tuning, contrastive alignment, or expert routing, including
TVL~\cite{fu2024tvl}, Touch100K~\cite{cheng2025touch100k},
Octopi~\cite{yu2024octopi}, and SToLa~\cite{cheng2026stola}. However, these
models are mainly trained with imitation or alignment objectives and do not
explicitly verify whether final answers are grounded in tactile evidence,
especially when visual and tactile cues conflict.

\noindent \textbf{Reinforcement learning for multimodal reasoning.}
DeepSeek-R1~\cite{guo2025deepseek} shows that GRPO~\cite{shao2024deepseekmath}
with rule-based rewards can elicit reasoning, inspiring extensions to visual
and video reasoning~\cite{huang2025vision,shen2025vlm,liu2025visual,feng2025video}
and more stable variants such as DAPO~\cite{yu2025dapo},
Dr.~GRPO~\cite{liu2025drgrpo}, and GSPO~\cite{zheng2025gspo}. Closely related
consistency-based methods include GRPO-CARE~\cite{chen2025grpo},
Visionary-R1~\cite{xia2025visionary}, and Video-R1's
T-GRPO~\cite{feng2025video}, which verify agreement across reasoning,
perception, answers, or temporal order. Touch-R1 extends this principle to
tactile reasoning by perturbing tactile inputs and rewarding policies that
remain sensitive to tactile evidence while enforcing ordinal-aware accuracy,
cross-sensor physical consistency, and structured reasoning.

\section{Method}
\vspace{-2mm}
\subsection{TouchReason-1M and TouchReason-Bench}
\label{sec:dataset
}

\begin{figure}[t]
\centering
\vspace{-7mm}
\resizebox{0.9\linewidth}{!}{\includegraphics{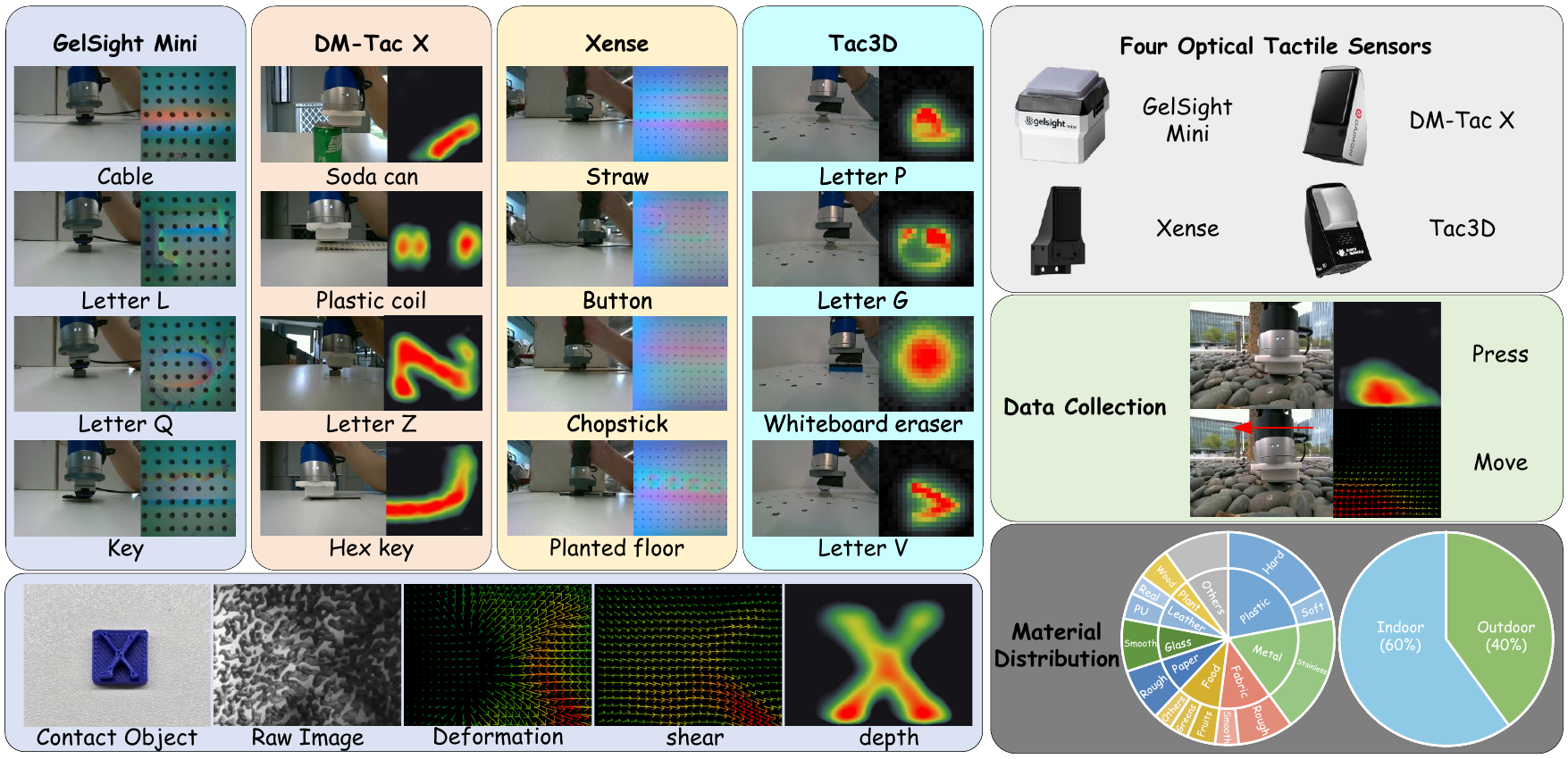}}
\vspace{-2mm}
\caption{Overview of TouchReason-1M collected with four optical tactile sensors on 1000+ objects across diverse materials and scenes.}
\label{fig:dataset}
\vspace{-4mm}
\end{figure} 

\textbf{Overview.}
We introduce TouchReason-1M, a multimodal tactile reasoning dataset containing over 1M synchronized tactile data pairs for R1-style training, together with TouchReason-Bench, a four-tier benchmark for evaluating tactile MLLMs. TouchReason-1M contains over 1,000 household and everyday objects across nine material categories: plastic, metal, fabric, food, paper, glass, leather, wood, and other materials. Each object is captured using four heterogeneous optical tactile sensors: GelSight Mini, Xense, Tac3D, and DM-Tac X. In this work, each tactile data pair refers to a synchronized tactile-force record, including the raw tactile observation, deformation field, shear field, depth field, force measurement, and associated metadata. Unlike prior tactile-language datasets that primarily target single-sensor alignment, tactile captioning, or property classification, TouchReason-1M is designed for tactile reasoning. It integrates cross-sensor tactile observations, force-conditioned contact dynamics, ordinal physical-property labels, and verifiable reasoning-style QA. This design supports both supervised tactile-language alignment and rule-based reinforcement learning, where final answers, structured outputs, and tactile-input dependence can be explicitly evaluated or regularized. 

\textbf{Acquisition.}
 Optical tactile sensors perceive contact by using an internal camera to capture the visual changes of a soft elastomer as it deforms during contact. Depending on the sensor structure and imaging principle, these changes may appear as sparse marker displacements, dense pattern deformation, or variations in brightness, color, and reflectance caused by surface deformation. Through dedicated processing, raw visual observations are converted into surface marker or pattern displacement fields, which are then used to estimate shear fields, depth fields, and contact force fields.

Taking DM-Tac X as an example, Fig.~\ref{fig:dataset} shows representative tactile signals generated during pressing and sliding interactions, including the deformed raw image, deformation field, shear field, and depth field. The deformation field is obtained from the raw image and represents the two-dimensional displacement field of the surface marker pattern under contact. The shear and depth fields are further estimated from the deformation field, with the shear field capturing tangential contact-related responses and the depth field describing local geometry associated with normal indentation. Together, these fields provide complementary cues for subsequent tactile perception and modeling.

As shown in Fig.~\ref{fig:dataset}, TouchReason-1M is collected through a force-guided two-stage acquisition protocol inspired by prior tactile collection pipelines. Each object is first recorded with calibrated RGB-D cameras to obtain visual appearance, depth geometry, and a contact-region reference. The object is then explored by four optical tactile sensors under the same standardized interaction protocol. In the first stage, each sensor performs continuous normal pressing from initial contact to 20 N, producing force-indexed deformation trajectories that capture hardness, compliance, and pressure-dependent responses. In the second stage, each sensor performs six exploratory motions at target normal-force levels to capture shear responses, local texture variations, and protrusion-induced deformation. All tactile streams are synchronized with compensated 3D contact force signals and stored as lossless frame stacks rather than compressed videos, preserving marker displacement, fine-grained texture, and deformation patterns critical for tactile reasoning. After filtering failed contacts, unstable force traces, saturated responses, and invalid contact regions, TouchReason-1M retains 14{,}700 valid tactile sequences and 1{,}323{,}000 synchronized tactile data pairs.

\begin{table}[t]
\vspace{-5mm}
\centering
\small
\caption{Comparison of TouchReason-1M with representative tactile and tactile-language datasets. TouchReason-1M is the first to combine multi-sensor optical tactile acquisition, force-guided interaction, ordinal physical labels, and verifiable reasoning-style QA.}
\label{tab:dataset_comparison}
\setlength{\tabcolsep}{4pt}
\renewcommand{\arraystretch}{1.2}
\resizebox{0.9\linewidth}{!}{%
\begin{tabular}{l c c l c c c c}
\toprule
\multirow{2}{*}{\textbf{Dataset}} &
\multirow{2}{*}{\textbf{Objects}} &
\multirow{2}{*}{\textbf{Touches / Frames}} &
\multirow{2}{*}{\textbf{Source}} &
\multirow{2}{*}{\textbf{Sensors}} &
\textbf{Force} &
\textbf{Ordinal} &
\textbf{Reason.} \\
& & & & &
\textbf{guided} &
\textbf{label} &
\textbf{QA} \\
\midrule
More Than a Feeling~\citep{yuan2017gelsight}       & 65            & 6.5K frames                    & Robot                           & GelSight          & \xmark  & \xmark  & \xmark \\
The Feeling of Success~\citep{calandra2017feeling} & 106           & 9.3K frames                    & Robot                           & GelSight          & \xmark  & \xmark  & \xmark \\
VisGel~\citep{li2019connecting}                    & 195           & 12K frames                     & Robot                           & GelSight          & \xmark  & \xmark  & \xmark \\
Touch and Go~\citep{yang2022touch}                 & 3{,}971       & 13.9K frames                   & Human                           & GelSight          & \xmark  & \xmark  & \xmark \\
TVL~\citep{fu2024tvl}                              & in-the-wild   & 44K pairs                      & Human + VLM                     & DIGIT             & \xmark  & \xmark  & \xmark \\
Touch100K~\citep{cheng2025touch100k}               & --            & 1M pairs                       & VLM + filter                    & GelSight          & \xmark  & \xmark  & \xmark \\
Octopi/PhysiCLeAR~\citep{yu2024octopi}             & 74            & 408 videos                     & Human                           & GelSight Mini     & partial & \cmark  & \xmark \\
AnyTouch/TacQuad~\citep{feng2025anytouch}          & --            & 72K frames                     & VLM                             & 4 sensors         & \xmark  & \xmark  & \xmark \\
\midrule
\rowcolor{gray!12}
\textbf{TouchReason-1M (Ours)}                     & \textbf{1000+} & \textbf{14K seq.\ / 1.32M fr.} & \textbf{Human + Gemini 2.5 Pro} & \textbf{4 sensors}& \cmark  & \cmark  & \cmark \\
\bottomrule
\end{tabular}}
\vspace{-6mm}
\end{table}

\textbf{Human Labeling.}
Two trained annotators assigned hardness, roughness, protrusion, and material labels for each object using synchronized RGB-D images, tactile sequences, and compensated 3D contact force signals. The first three attributes were encoded as three-level ordinal attributes and cross-verified against force--deformation curves, shear responses, and contact geometry. Inter-annotator Cohen's $\kappa$ was 0.82, 0.78, and 0.80 for the three ordinal attributes, with 94.3\% exact-match agreement for material labels; disagreements were adjudicated by a senior annotator.

\begin{figure}[t]
\vspace{-5mm}
\centering
\resizebox{\linewidth}{!}{\includegraphics{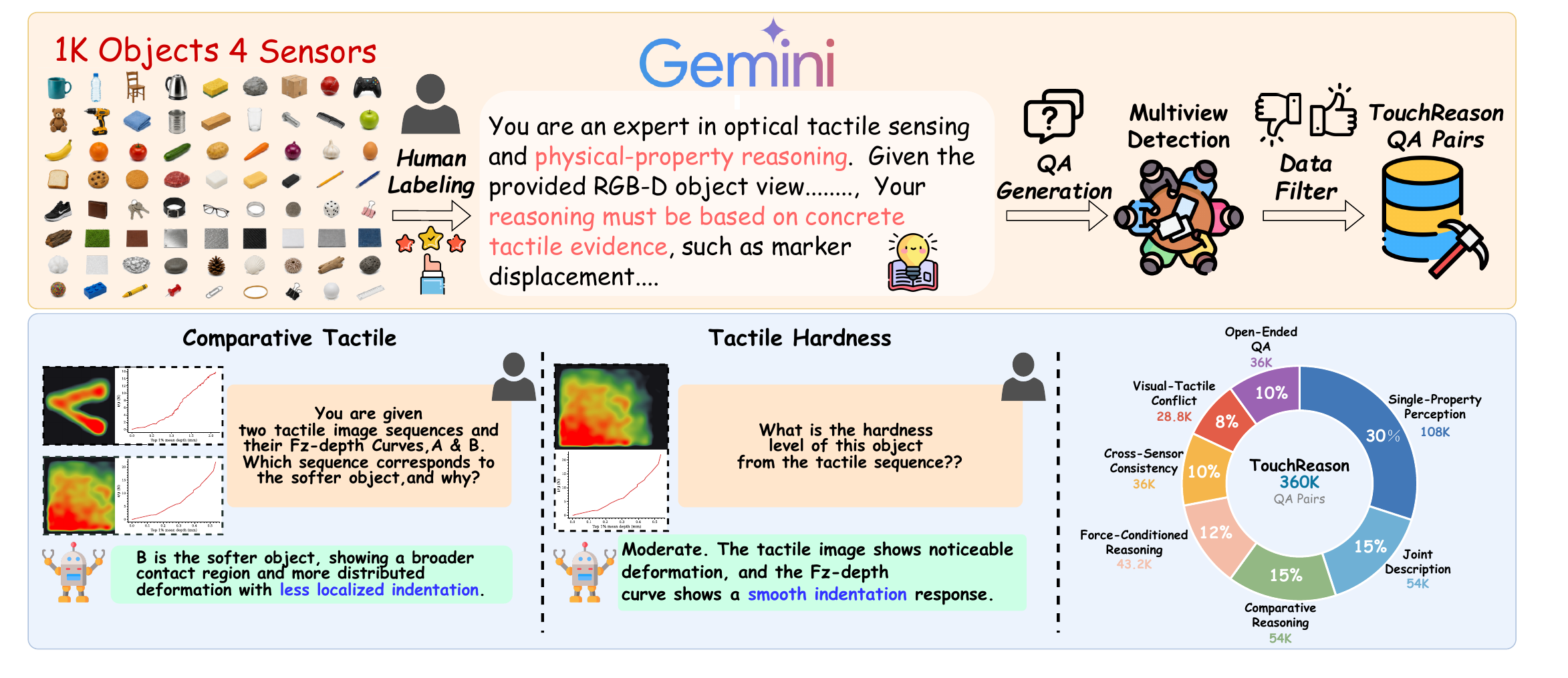}}
\vspace{-5mm}
\caption{Overview of TouchReason QA Pairs construction. The upper panel shows data collection, human labeling, QA generation, multiview detection, and filtering pipeline. The lower panel displays typical QA examples for comparative tactile reasoning and hardness recognition, together with the distribution of QA task types.}
\label{fig:QA_generate}
\vspace{-1.6em}
\end{figure}

\textbf{QA Generation and Quality Verification.}
As shown in Fig.~\ref{fig:QA_generate}, based on the human-labeled tactile data pairs, Gemini 2.5 Pro was used to generate candidate reasoning-style QA pairs and rationales from RGB-D views, tactile frame stacks, force traces, sensor metadata, and contact ROIs. The prompt explicitly required each rationale to ground its conclusion in concrete tactile evidence, such as marker displacement, deformation magnitude, surface texture, or force-dependent response, rather than vague assertions. We randomly sampled 2{,}000 generated pairs for independent human evaluation, yielding a 91.6\% pass rate across three axes: tactile-evidence grounding, consistency with human ordinal labels, and self-consistency of the reasoning chain. Rule-based filters further removed label description conflicts, hallucinated contact regions, missing answers, and reasoning--answer inconsistencies.

 \subsection{Touch-R1}
\label{sec:method}
\begin{figure*}[t]
\vspace{-5mm}
\centering
\makebox[\textwidth][c]{
\includegraphics[width=1.1\textwidth]{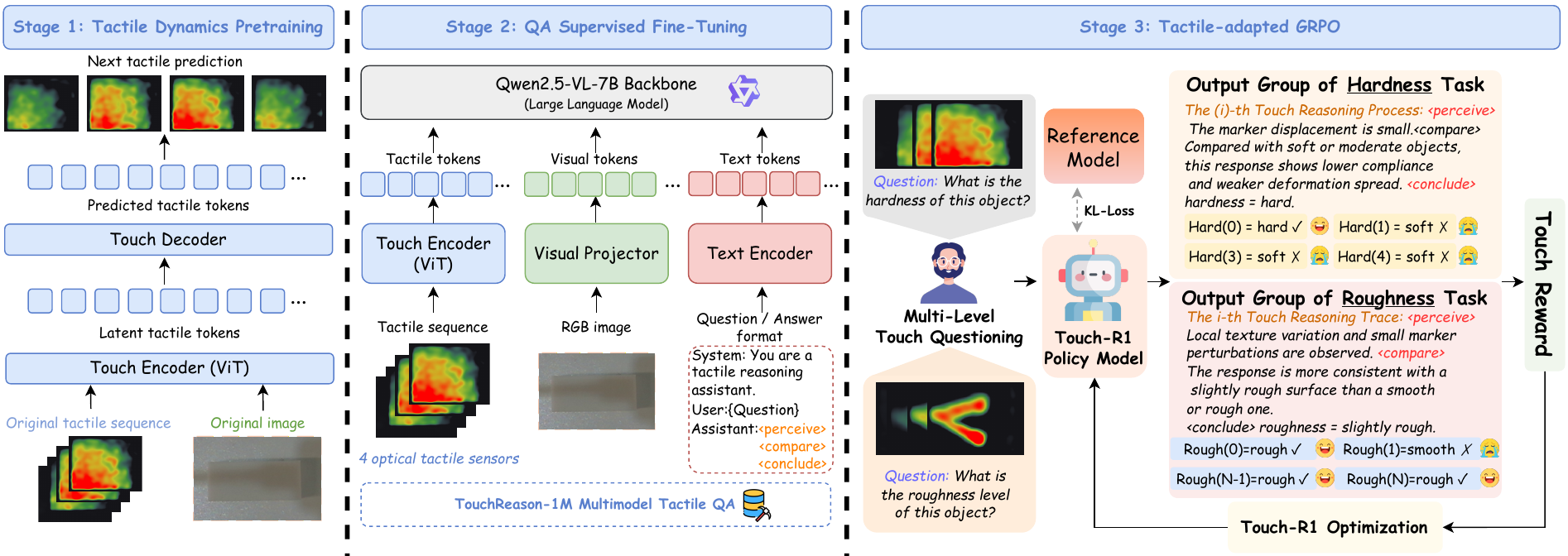}
}
\vspace{-5mm}
\caption{\textbf{Touch-R1: a three-stage framework for tactile-grounded reasoning.}
\textbf{(1) Tactile Dynamics Pretraining}: a ViT touch encoder is pretrained by predicting future tactile tokens from past ones, capturing deformation dynamics across optical tactile sensors.
\textbf{(2) QA Supervised Fine-Tuning}: tactile, visual, and text tokens are aligned to a Qwen2.5-VL-7B backbone on TouchReason-1M, supervising the assistant to answer under a structured \texttt{<perceive>}–\texttt{<compare>}–\texttt{<conclude>} template.
\textbf{(3) Tactile-adapted GRPO}: starting from the SFT checkpoint, rollouts are scored by an ordinal-aware accuracy reward, a cross-sensor physical consistency reward, and a structured-format reward, optimizing the policy toward answers that are correct, cross-sensor consistent, and grounded in tactile evidence.}
 
\label{fig:overview}
\vspace{-1em}
\end{figure*}

This section introduces Touch-R1, a reinforcement-learning framework for
tactile-grounded multi-sensor reasoning, as shown in Fig.~\ref{fig:overview}.
Our key observation is that answer correctness alone is insufficient for tactile reasoning: a policy may predict the correct label while relying on category
priors, ignoring tactile streams, or producing cross-sensor contradictions.
Such answer-only success masks poor grounding, where the output is correct but
the reasoning may not be physically reliable.

Touch-R1 addresses this issue by augmenting GRPO with three output-side rewards
and one input-side grounding objective. The output-side rewards capture ordinal
tactile attributes, enforce cross-sensor physical consistency, and stabilize
structured reasoning, while the input-side objective perturbs tactile streams
to encourage policy sensitivity to tactile evidence. As a result, Touch-R1
optimizes for answers that are not only correct, but also grounded in
physically consistent and input-dependent tactile evidence.

\noindent \textbf{Ordinal-aware reward.}
Tactile attributes such as hardness, roughness, and protrusion are not purely categorical; when elasticity is annotated, it is also treated as an ordered physical attribute. A binary correctness reward ignores this structure and treats adjacent mistakes the same as distant mistakes. For example, if the ground-truth hardness level is high, predicting a neighboring level is physically closer to the truth than predicting the opposite end of the scale. Under sparse answer-only feedback, this mismatch can encourage the policy to collapse toward frequent middle levels, leading to ordinal collapse.

To counteract this failure, we replace binary correctness with an ordinal-aware reward. Let $y\in\{0,\ldots,K-1\}$ denote the ground-truth ordinal label, let $\hat{y}$ denote the prediction parsed from the \texttt{<conclude>} segment, and let $d=|\hat{y}-y|$ be the ordinal distance. We define
\begin{equation}
R_{\mathrm{acc}}(\hat{y},y)=
\begin{cases}
1.0, & \hat{y}\ \text{is parsable and } d=0,\\
0.4, & \hat{y}\ \text{is parsable and } d=1,\\
0.0, & \hat{y}\ \text{is parsable and } d\ge 2,\\
-0.1, & \hat{y}\ \text{is unparsable}.
\end{cases}
\label{eq:racc}
\end{equation}
This reward encodes the prior that predictions closer on the ordinal axis are closer in physical meaning. The adjacent-error reward remains strictly below the exact-match reward, so it does not weaken the incentive for correct prediction; however, it is still preferred over distant errors, which densifies the reward signal along the answer space. For categorical labels such as material, we use the exact-match variant of this reward, assigning 1.0 to correct parsable predictions and 0 otherwise.

\noindent \textbf{Cross-sensor physical consistency reward.}
Answer-only rewards also fail to penalize cross-sensor contradictions. For the same object, heterogeneous tactile sensors such as GelSight Mini, Xense, Tac3D, and DM-Tac X may produce different low-level signals, but their summarized physical attributes should agree after being mapped into the same ordinal space. If the policy reports conflicting per-sensor evidence but happens to predict the correct final label, an answer-only reward will still reinforce the rollout.

We therefore introduce a cross-sensor physical consistency reward. Let $p_a$ and $p_b$ denote the ordinal property summaries parsed from the \texttt{<perceive>} segment for two sensors $a$ and $b$, and let $p_c$ denote the cross-sensor conclusion parsed from the \texttt{<compare>} segment. The normalized ordinal gap is $w(u,v)=\min(|u-v|/(K-1),1)$. We define
\begin{equation}
R_{\mathrm{cs}}=
\begin{cases}
1-\frac{1}{2}\big(w(p_a,p_c)+w(p_b,p_c)\big),
& p_a,p_b,p_c\ \text{are parsable},\\
0, & \text{otherwise}.
\end{cases}
\label{eq:rcs}
\end{equation}
This reward reaches its maximum when both per-sensor summaries agree with the cross-sensor conclusion. Small ordinal disagreements incur mild penalties, whereas distant contradictions are penalized more strongly. The reward does not require different sensors to produce identical raw tactile signals; instead, it requires their summarized physical-property evidence to be consistent. This encourages the policy to ground its conclusion in shared tactile evidence rather than sensor-specific artifacts or language shortcuts.

\noindent \textbf{Structured-format reward.}
Touch-R1 uses a three-segment reasoning template with \texttt{<perceive>}, \texttt{<compare>}, and \texttt{<conclude>} segments. This template is not merely cosmetic: it defines the interface through which the reward functions parse per-sensor evidence, cross-sensor comparison, and the final ordinal prediction. Once RL training causes the output format to drift, $R_{\mathrm{acc}}$ and $R_{\mathrm{cs}}$ become unreliable because the required fields may be missing, duplicated, disordered, or unparsable.

To stabilize this interface, we use a structured-format reward:
\begin{equation}
R_{\mathrm{fmt}}(o)=
\mathbf{1}\!\left[
o\ \text{matches}\ 
\langle\mathrm{perceive}\rangle\cdots
\langle\mathrm{compare}\rangle\cdots
\langle\mathrm{conclude}\rangle
\right].
\label{eq:rfmt}
\end{equation}
The deterministic parser requires all opening and closing tags to appear exactly once and in order. We keep this reward binary rather than continuous because partially formatted outputs are already penalized by the accuracy and consistency rewards. Assigning extra reward to partial formats can introduce redundant signals and encourage superficially valid but semantically empty templates.

The scalar reward used by GRPO combines the three output-side rewards:
\begin{equation}
R=
\lambda_{\mathrm{acc}}R_{\mathrm{acc}}
+\lambda_{\mathrm{cs}}R_{\mathrm{cs}}
+\lambda_{\mathrm{fmt}}R_{\mathrm{fmt}},
\qquad
(\lambda_{\mathrm{acc}},\lambda_{\mathrm{cs}},\lambda_{\mathrm{fmt}})
=(0.6,0.3,0.1).
\label{eq:rtotal}
\end{equation}
Here $R_{\mathrm{acc}}$ provides the primary supervision for ordinal prediction, $R_{\mathrm{cs}}$ regularizes cross-sensor physical consistency, and $R_{\mathrm{fmt}}$ preserves a parsable reasoning interface.

\noindent \textbf{Implicit tactile grounding objective.}
The three rewards above supervise what the policy outputs, but they do not guarantee the policy actually uses tactile input. A policy can exploit object-category priors, visual appearance, or language shortcuts to predict tactile attributes, remaining nearly unchanged even when a tactile stream is corrupted. We refer to this behavior as a tactile-blind shortcut.

To address this issue, we introduce an implicit tactile grounding objective. Let $T$ denote the original tactile observation, and let $T_{\mathrm{mask}}$ denote a perturbed observation obtained by replacing one randomly selected tactile stream with shape-matched Gaussian noise. Given question $q$ and prefix $y_{<t}$, we compare the policy distributions under the intact and perturbed tactile inputs:
\begin{equation}
G_{\mathrm{tg}}(\theta)=
\mathbb{E}_{t}\!\left[
D_{\mathrm{KL}}\!\left(
\pi_\theta(\cdot\mid q,T,y_{<t})
\;\Vert\;
\pi_\theta(\cdot\mid q,T_{\mathrm{mask}},y_{<t})
\right)
\right].
\label{eq:gtg}
\end{equation}
 In practice, we estimate the expectation by averaging over generated token
positions. Maximizing $G_{\mathrm{tg}}$ encourages output distributions to
shift under tactile perturbations, yielding a label-free signal for input
dependence without explicit perception targets.

$R_{\mathrm{cs}}$ and $G_{\mathrm{tg}}$ impose complementary constraints. The consistency reward checks whether the generated per-sensor evidence agrees in physical-property space, while the grounding objective checks whether the policy distribution depends on tactile observations themselves. The former regularizes output-side physical consistency; the latter enforces input-side tactile dependence.

\noindent \textbf{Overall objective.}
We optimize the policy with standard GRPO using the composite reward $R$, and add the tactile grounding objective as an input-dependence regularizer. The final Touch-R1 objective is
\begin{equation}
\mathcal{J}_{\mathrm{Touch\text{-}R1}}(\theta)
=
\mathcal{J}_{\mathrm{GRPO}}(\theta;R)
+
\gamma G_{\mathrm{tg}}(\theta),
\label{eq:touchr1_obj}
\end{equation}
where $\gamma$ controls the strength of tactile grounding.

\begin{table}[t]
\vspace{-5mm}
\centering
\scriptsize
\setlength{\tabcolsep}{4.2pt}
\renewcommand{\arraystretch}{1.05}
\caption{Main results on TouchReason-Bench. OMAE is lower-better; all other metrics are higher-better. Avg denotes the aggregate score over the full metric set, as defined in Appendix~B.2.}
\label{tab:main}
\resizebox{\linewidth}{!}{%
\begin{tabular}{@{}lcccccccccc@{}}
\toprule
\multirow{2}{*}{Model}
& \multicolumn{4}{c}{Property identification}
& \multicolumn{2}{c}{Ordinal reasoning}
& \multicolumn{3}{c}{Cross-sensor consistency}
& \multirow{2}{*}{Avg} \\
\cmidrule(lr){2-5}\cmidrule(lr){6-7}\cmidrule(lr){8-10}
& H-Acc & R-Acc & P-Acc & Mat-Acc
& OMAE$\downarrow$ & L2-EM
& SFD & SOI-$\rho$ & CSC
& \\
\midrule

\rowcolor{GroupGray}
\multicolumn{11}{c}{\textsc{Closed-source Frontier MLLMs}} \\
GPT-4o~\citep{gpt4o}             & 48.2 & 41.5 & 39.7 & 58.4 & 0.71 & 11.3 & 53.0 & 0.18 & 39.6 & 35.4 \\
Gemini-2.5-Pro~\citep{gemini2.5}     & 51.1 & 44.7 & 42.3 & 60.2 & 0.67 & 13.8 & 56.4 & 0.22 & 41.8 & 37.6 \\
Claude-3.5-Sonnet~\citep{Claude-3.5}  & 46.3 & 39.1 & 37.5 & 55.8 & 0.74 &  9.8 & 50.7 & 0.15 & 36.9 & 33.5 \\

\addlinespace[3pt]
\rowcolor{GroupGray}
\multicolumn{11}{c}{\textsc{Open-source General VLMs}} \\
LLaVA-OneVision-7B~\citep{LLaVA-OneVision} & 37.6 & 32.4 & 30.1 & 45.8 & 0.86 &  5.7 & 45.9 & 0.08 & 30.2 & 26.9 \\
Qwen2.5-VL-7B~\citep{Qwen2.5}      & 39.4 & 33.7 & 31.8 & 47.2 & 0.83 &  6.2 & 47.1 & 0.09 & 31.4 & 28.0 \\
InternVL2.5-8B~\citep{Intern-VL2.5}     & 40.8 & 35.2 & 33.0 & 49.6 & 0.81 &  7.5 & 48.6 & 0.11 & 33.1 & 29.4 \\

\addlinespace[3pt]
\rowcolor{GroupGray}
\multicolumn{11}{c}{\textsc{Tactile-specialist Models}} \\
Octopi-7B~\citep{yu2024octopi}          & 53.4 & 45.9 & 43.1 & 51.7 & 0.62 & 14.5 & 58.2 & 0.27 & 38.4 & 38.1 \\
Octopi-13B~\citep{yu2024octopi}         & 58.1 & 50.3 & 47.6 & 54.9 & 0.55 & 17.9 & 62.5 & 0.34 & 41.2 & 41.7 \\
AnyTouch+Qwen~\citep{feng2025anytouch,Qwen2.5}      & 60.2 & 52.7 & 50.4 & 62.1 & 0.51 & 19.3 & 64.1 & 0.39 & 49.7 & 45.0 \\
VTV-LLM-7B~\citep{VTV-LLM}         & 61.5 & 54.1 & 51.8 & 63.0 & 0.49 & 20.2 & 65.4 & 0.41 & 50.6 & 46.3 \\
SToLa~\citep{cheng2026stola}              & 63.7 & 57.4 & 54.6 & 64.3 & 0.45 & 22.1 & 66.8 & 0.43 & 51.9 & 47.5 \\

\midrule
\rowcolor{OursGray}
Touch-R1-7B        & \best{76.8} & \best{68.4} & \best{65.1} & \best{74.6} & \best{0.24} & \best{39.5} & \best{78.2} & \best{0.66} & \best{71.3} & \best{60.1} \\
\bottomrule
\end{tabular}}
\vspace{-5mm}
\end{table}

\begin{table}[t]
\caption{(a) Cumulative effect of each Touch-R1 component, starting from the Qwen2.5-VL-7B base. (b) Per-attribute Tier-1 accuracy across hardness, roughness, protrusion, and material.}
\label{tab:ablation_combined}
\centering
\small
\begin{minipage}[t]{0.44\linewidth}
\centering
\textbf{(a) Cumulative additions}\\[2pt]
\resizebox{0.8\linewidth}{!}{%
\begin{tabular}{lccc}
\toprule
Configuration & OMAE$\downarrow$ & CSC & Avg \\
\midrule
Qwen2.5-VL-7B (zero-shot)   & 0.83 & 31.4 & 28.0 \\
\,$+$ Cold-start SFT        & 0.46 & 52.6 & 47.8 \\
\,$+$ GRPO (0/1)            & 0.42 & 54.3 & 50.1 \\
\,$+$ Ordinal Reward        & 0.31 & 56.1 & 54.6 \\
\,$+$ Consistency Reward    & 0.27 & 69.2 & 58.3 \\
\,$+$ Format Reward         & 0.26 & 70.4 & 59.2 \\
\rowcolor{gray!22}\textbf{Touch-R1} & \textbf{0.24} & \textbf{71.3} & \textbf{60.1} \\
\bottomrule
\end{tabular}}
\end{minipage}
\hfill
\begin{minipage}[t]{0.55\linewidth}
\centering
\textbf{(b) Per-attribute accuracy (Tier-1)}\\[2pt]
\resizebox{\linewidth}{!}{%
\begin{tabular}{lccccc}
\toprule
Method & Hardness & Roughness & Protrusion & Material & Avg \\
\midrule
Qwen2.5-VL-7B          & 31.2 & 28.9 & 25.7 & 26.1 & 28.0 \\
\,$+$ Cold-start SFT   & 49.4 & 47.2 & 45.8 & 48.7 & 47.8 \\
\,$+$ GRPO (0/1)       & 51.7 & 49.8 & 48.3 & 50.6 & 50.1 \\
\,$+$ Ordinal Reward   & 56.3 & 54.1 & 52.7 & 55.4 & 54.6 \\
\,$+$ Consistency      & 60.2 & 57.8 & 55.9 & 59.4 & 58.3 \\
\rowcolor{gray!22}\textbf{Touch-R1} & \textbf{62.4} & \textbf{59.7} & \textbf{57.6} & \textbf{60.7} & \textbf{60.1} \\
\bottomrule
\end{tabular}}
\end{minipage}
\vspace{-3mm}
\end{table}

\section{Experiments}
\label{sec:experiments}

\subsection{Experimental Setup}
\label{sec:setup}

\noindent \textbf{Training data and protocol.}
We train Touch-R1 on TouchReason-1M and evaluate it primarily on TouchReason-Bench. Training follows a three-stage protocol. First, we pretrain the tactile backbone on 1M sampled tactile frames from the training split, where 1M denotes frame-level tactile samples rather than interaction sequences, using future tactile-frame prediction to encourage the model to capture force-conditioned contact dynamics. Second, we perform cold-start SFT on a curated 240K subset of generated and verified QA pairs to initialize the structured \texttt{<perceive>} \texttt{<compare>} \texttt{<conclude>} reasoning format. Third, we apply tactile-grounded GRPO on the remaining 120K generated QA samples, where final conclusions are evaluated by rule-based tactile rewards. The three stages use disjoint training subsets to avoid leakage. Unless otherwise specified, the language backbone is Qwen2.5-VL-7B-Instruct. All experiments are conducted on 16 NVIDIA H200 GPUs.

\noindent \textbf{Benchmarks.}
TouchReason-Bench contains 4,800 QA pairs over 200 held-out objects that are unseen during training. It covers four tactile sensor types, GelSight Mini, Xense, Tac3D, and DM-Tac X, and nine material categories. To evaluate generalization beyond our benchmark, we additionally report tactile video prediction on Touch-and-Go and TouchReason-Video, and compare with the tactile-specialized VTV-LLM-7B on VTV150K. We compare Touch-R1 with three groups of baselines: closed-source frontier MLLMs, including GPT-4o, Gemini-2.5-Pro, and Claude-3.5-Sonnet; open-source general-purpose VLMs, including Qwen2.5-VL-7B, InternVL2.5-8B, and LLaVA-OneVision-7B; and tactile-specialist models, including Octopi, AnyTouch+Qwen, SToLa, and VTV-LLM-7B.

\noindent \textbf{Implementation details.}
Each tactile sequence is downsampled to 8 frames at $448{\times}448$ resolution. GRPO uses group size $G{=}8$ and KL coefficient $\beta{=}10^{-3}$. We use AdamW with learning rates $1{\times}10^{-4}$ for tactile pretraining, $1{\times}10^{-5}$ for cold-start SFT, and $1{\times}10^{-6}$ for GRPO. Additional hyperparameters are provided in Appendix~B.1.

\noindent \textbf{Metrics.}
 TouchReason-Bench evaluates four aspects of tactile reasoning. Property identification is measured by H-Acc, R-Acc, P-Acc, and Mat-Acc, corresponding to hardness, roughness, protrusion, and material classification. Ordinal and compositional reasoning are measured by OMAE and L2-EM, where OMAE is the mean absolute ordinal error and L2-EM is the exact-match rate for multi-attribute answers. Comparative reasoning is measured by SFD and SOI-$\rho$, which evaluate pairwise and listwise tactile comparisons. Cross-sensor consistency is measured by CSC. OMAE is lower-better, and all other metrics are higher-better. Formal metric definitions are given in Appendix~B.2.

\begin{figure}[t]
\vspace{-5mm}
    \centering
    \includegraphics[width=0.8\linewidth]{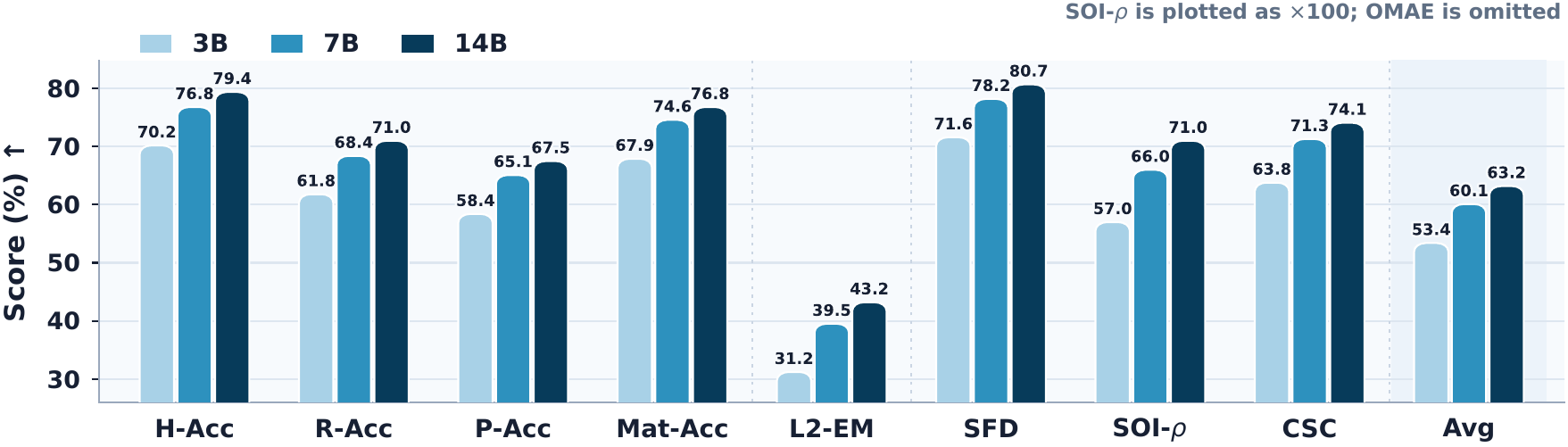}
    \vspace{-2mm}
    \caption{Scaling behavior of Touch-R1 across 3B, 7B, and 14B backbones on TouchReason-Bench. SOI-$\rho$ is plotted as $\times 100$; OMAE is omitted because it is lower-better.}
    \label{fig:scale_task_bars}
    \vspace{-1.7em}
\end{figure}

\begin{table}[t]
\centering
\scriptsize
\setlength{\tabcolsep}{3.2pt}
\renewcommand{\arraystretch}{0.96}
\caption{Tactile video prediction results on Touch-and-Go and TouchReason-Video. }
\vspace{-0.3mm}
\label{tab:video_prediction}
\resizebox{0.8\linewidth}{!}{%
\begin{tabular}{lllcccccc}
\toprule
\multirow{2}{*}{Horizon}
& \multirow{2}{*}{Method}
& \multirow{2}{*}{Modality}
& \multicolumn{3}{c}{Touch-and-Go}
& \multicolumn{3}{c}{TouchReason-Video} \\
\cmidrule(lr){4-6} \cmidrule(lr){7-9}
& &
& L1$\downarrow$ & SSIM$\uparrow$ & LPIPS$\downarrow$
& L1$\downarrow$ & SSIM$\uparrow$ & LPIPS$\downarrow$ \\
\midrule
\multirow{6}{*}{Skip 3}
& SVG~\citep{denton2018stochastic}     & Touch only       & 0.782 & 0.572 & 0.391 & 0.834 & 0.529 & 0.428 \\
& Geng et al.~\citep{geng2022touch}    & Touch only       & 0.628 & 0.708 & 0.103 & 0.681 & 0.661 & 0.146 \\
& Ours                                 & Touch only       & \textbf{0.574} & \textbf{0.742} & \textbf{0.082} & \textbf{0.627} & \textbf{0.696} & \textbf{0.118} \\
\cmidrule(lr){2-9}
& SVG~\citep{denton2018stochastic}     & Touch + vision   & 0.757 & 0.602 & 0.368 & 0.812 & 0.553 & 0.409 \\
& Geng et al.~\citep{geng2022touch}    & Touch + vision   & 0.617 & 0.719 & 0.091 & 0.668 & 0.674 & 0.132 \\
& Ours                                 & Touch + vision   & \textbf{0.552} & \textbf{0.763} & \textbf{0.071} & \textbf{0.608} & \textbf{0.715} & \textbf{0.106} \\
\midrule
\multirow{6}{*}{Skip 5}
& SVG~\citep{denton2018stochastic}     & Touch only       & 0.807 & 0.513 & 0.412 & 0.861 & 0.468 & 0.454 \\
& Geng et al.~\citep{geng2022touch}    & Touch only       & 0.691 & 0.698 & 0.279 & 0.746 & 0.648 & 0.331 \\
& Ours                                 & Touch only       & \textbf{0.638} & \textbf{0.724} & \textbf{0.241} & \textbf{0.694} & \textbf{0.676} & \textbf{0.291} \\
\cmidrule(lr){2-9}
& SVG~\citep{denton2018stochastic}     & Touch + vision   & 0.762 & 0.546 & 0.397 & 0.819 & 0.501 & 0.441 \\
& Geng et al.~\citep{geng2022touch}    & Touch + vision   & 0.663 & 0.713 & 0.265 & 0.721 & 0.663 & 0.314 \\
& Ours                                 & Touch + vision   & \textbf{0.612} & \textbf{0.742} & \textbf{0.228} & \textbf{0.671} & \textbf{0.692} & \textbf{0.277} \\
\bottomrule
\end{tabular}}
\vspace{-1.6em}
\end{table}

\subsection{Main Results on TouchReason-Bench}
\label{sec:main_results}

Table~\ref{tab:main} shows that Touch-R1-7B achieves an Avg score of 60.1, outperforming the strongest tactile-specialist baseline, SToLa, by 12.6 percentage points and the strongest closed-source baseline, Gemini-2.5-Pro, by 22.5 percentage points. The gains are largest on metrics that require structured tactile reasoning: Touch-R1 improves CSC by 19.4 percentage points and L2-EM by 17.4 percentage points over SToLa. OMAE also decreases from 0.45 to 0.24, indicating that Touch-R1 reduces the ordinal distance of incorrect predictions from the ground truth.

\vspace{-1em}
\subsection{Ablation and Analysis}
\label{sec:ablation}

 Table~\ref{tab:ablation_combined} highlights how each component contributes to Touch-R1. Cold-start SFT provides the strongest initial boost, raising Avg from 28.0 to 47.8 by teaching the model the structured tactile reasoning format. Adding standard GRPO improves Avg to 50.1, but the gain remains limited, suggesting that final-answer supervision alone cannot ensure reliable tactile-grounded reasoning. The Touch-R1 components address distinct failure modes. The ordinal reward lowers OMAE from 0.42 to 0.31, indicating that distance-aware feedback mitigates ordinal collapse. The consistency reward raises CSC from 56.1 to 69.2, giving the largest improvement on cross-sensor physical consistency. The format reward brings a smaller but stable gain by preserving the parsable reasoning interface. The per-attribute breakdown further shows that these improvements are not confined to one property, but spread across hardness, roughness, protrusion, and material recognition.

\noindent \textbf{Scale analysis.} Figure~\ref{fig:scale_task_bars} reports the task-wise scaling behavior of Touch-R1. Performance improves consistently with model size on most higher-better metrics, with especially clear gains on L2-EM, SOI-$\rho$, CSC, and Avg. This suggests that larger backbones strengthen cross-attribute reasoning and cross-sensor consistency, while the Touch-R1 objective remains effective across scales.

\begin{figure}[t]
\vspace{-5mm}
    \centering
\resizebox{0.83\linewidth}{!}{\includegraphics{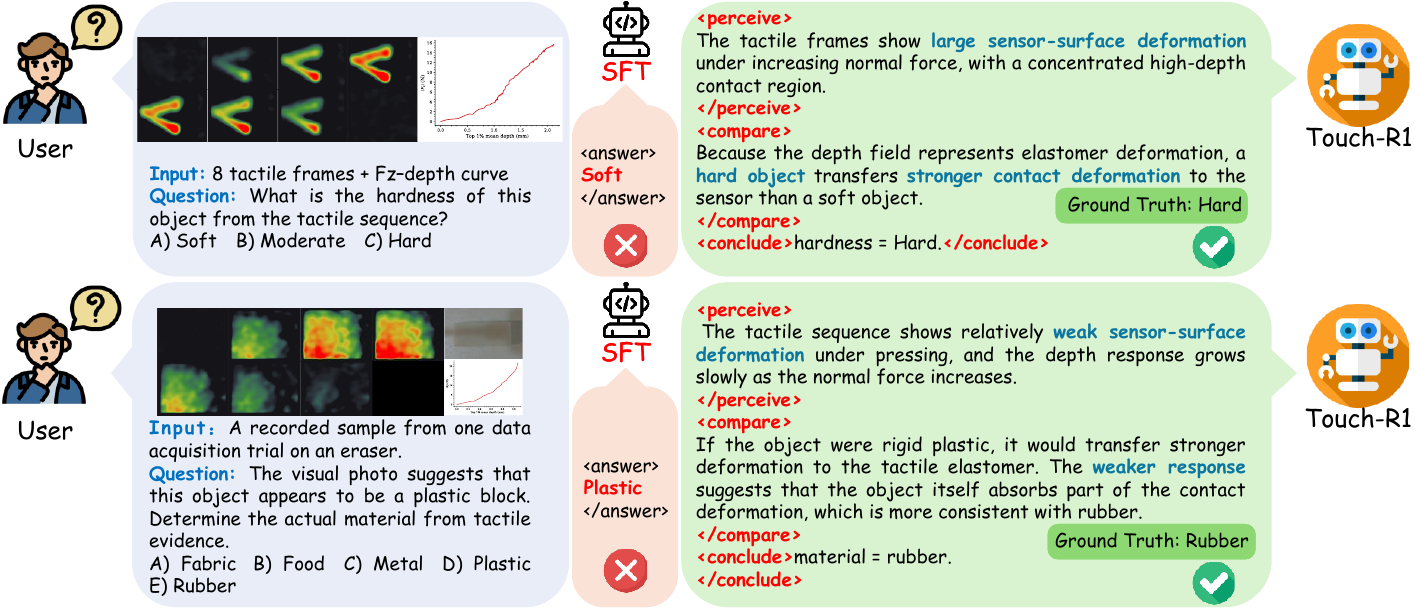}}
\vspace{-0.5em}
\captionof{figure}{Qualitative examples of Touch-R1 reasoning. The model uses tactile evidence to estimate hardness and resolves visual-tactile conflicts in material recognition.}
\label{fig:qualitative}
\vspace{-0.5em}
\end{figure}




\begin{table}[t]
\centering
\vspace{-0.2em}
\scriptsize
\setlength{\tabcolsep}{3.2pt}
\renewcommand{\arraystretch}{0.92}
\captionof{table}{Comparison on VTV150K. Results are reported in percentages except SOI, following the official evaluation protocol.}
\label{tab:vtv150k_comparison}
\vspace{-0.2em}
\resizebox{0.82\linewidth}{!}{%
\begin{tabular}{lcccccccccc}
\toprule
Model & Hardness & Protrusion & Elasticity & Friction & Combined & SFD & SOI & OSC & TSA & Average \\
\midrule
GPT-4o~\citep{gpt4o}               & 34.7 & 32.6 & 32.6 & 18.7 &  2.1 & 40.9 & 38.4 & 16.6 & 36.0 & 28.0 \\
Gemini-2.5-Pro-Exp~\citep{gemini2.5}   & 36.2 & 34.7 & 39.1 & 21.0 &  4.3 & 42.6 & 29.4 & 18.5 & 40.0 & 29.5 \\
LLaVA-OneVision-7B~\citep{LLaVA-OneVision}   & 27.5 & 32.6 & 26.0 & 20.2 &  0.7 & 40.9 & 28.2 & 11.7 & 30.0 & 24.2 \\
LLaVA-Video-Qwen2-7B~\citep{LLaVA-Video} & 30.4 & 29.7 & 28.9 & 18.1 &  2.1 & 33.6 & 29.4 & 17.2 & 36.0 & 25.0 \\
InternVL2.5-VL-8B~\citep{Intern-VL2.5}    & 18.1 & 23.9 & 21.0 & 13.7 &  0.0 & 24.5 & 17.9 & 11.1 & 24.0 & 17.1 \\
VideoLLaMA3-7B~\citep{VideoLLa}       & 15.2 & 21.7 & 14.4 & 10.8 &  0.0 & 11.4 & 12.8 &  7.4 & 20.0 & 12.6 \\
Qwen2.5-VL-7B~\citep{Qwen2.5}        & 25.3 & 28.9 & 17.3 & 15.9 &  1.4 & 22.9 & 28.2 & 16.0 & 30.0 & 20.6 \\
VTV-LLM-7B~\citep{VTV-LLM}           & 73.9 & 75.0 & 67.3 & 56.5 & 35.6 & 71.3 & 57.6 & 43.2 & 64.0 & 60.4 \\
\rowcolor{gray!10}
Ours                 & \textbf{78.6} & \textbf{79.3} & \textbf{71.8} & \textbf{61.2} & \textbf{41.9} & \textbf{76.5} & \textbf{63.4} & \textbf{48.7} & \textbf{68.0} & \textbf{65.5} \\
\bottomrule
\end{tabular}}
\vspace{-0.9em}
\end{table}

\noindent \textbf{Qualitative Analysis.}
Figure~\ref{fig:qualitative} illustrates cases where visual appearance is insufficient or misleading. Touch-R1 grounds its conclusions in tactile cues such as marker displacement and deformation patterns, leading to physically plausible predictions.
\label{sec:qualitative}

\FloatBarrier

\vspace{-1em}
\section{Conclusion}
\vspace{-1em}
Touch-R1 introduces the R1 paradigm to heterogeneous optical tactile reasoning. By combining ordinal-aware rewards, cross-sensor physical consistency, structured-format control, and input-side tactile grounding, Touch-R1 improves tactile reasoning accuracy, reduces ordinal error, and strengthens cross-sensor consistency. Experiments on TouchReason-Bench and external tactile-video benchmarks show that Touch-R1 can use physical contact evidence to revise misleading visual priors and produce structured perception-comparison-revision rationales.

\bibliographystyle{plain}
\bibliography{main}




\end{document}